\documentclass[11pt,a4paper]{article}
\usepackage[hyperref]{emnlp2020}
\usepackage{times}
\usepackage{latexsym}

\usepackage{url}

\usepackage{microtype}

\usepackage{multirow}
\usepackage{graphicx}
\usepackage{CJKutf8}
\usepackage{tabularx}

\usepackage{makecell}

\aclfinalcopy 
\def\aclpaperid{468} 

\title{KorNLI and KorSTS: \\New Benchmark Datasets for Korean Natural Language Understanding}

\author{Jiyeon Ham\thanks{\ \ Equal Contribution.} , Yo Joong Choe\footnotemark[1] , Kyubyong Park\footnotemark[1] , Ilji Choi, Hyungjoon Soh \\
  Kakao Brain \\
  {\small\tt \{jiyeon.ham,yj.choe,kyubyong.park,ilji.choi,hj.soh\}@kakaobrain.com} \\
  }

\date{}

\begin{document}
\maketitle
\begin{abstract}
Natural language inference (NLI) and semantic textual similarity (STS) are key tasks in natural language understanding (NLU). 
 Although several benchmark datasets for those tasks have been released in English and a few other languages, there are no publicly available NLI or STS datasets in the Korean language.
 Motivated by this, we construct and release new datasets for Korean NLI and STS, dubbed KorNLI and KorSTS, respectively. 
 Following previous approaches, we machine-translate existing English training sets and manually translate development and test sets into Korean. 
 To accelerate research on Korean NLU, we also establish baselines on KorNLI and KorSTS.
 Our datasets are publicly available at \url{https://github.com/kakaobrain/KorNLUDatasets}.
\end{abstract}

\section{Introduction}
\label{intro}

Natural language inference (NLI) and semantic textual similarity (STS) are considered as two of the central tasks in natural language understanding (NLU).
They are not only featured in GLUE \citep{wang2018glue} and SuperGLUE \citep{wang2019superglue}, which are two popular benchmarks for NLU, but also known to be useful for supplementary training of pre-trained language models \citep{phang2018sentence} as well as for building and evaluating fixed-size sentence embeddings \citep{reimers2019sentencebert}.
Accordingly, several benchmark datasets have been released for both NLI \citep{bowman2015large,williams2018broad} and STS \citep{cer2017semeval} in the English language. 

When it comes to the Korean language, however, benchmark datasets for NLI and STS do not exist.
Popular benchmark datasets for Korean NLU typically involve question answering\footnote{\url{https://korquad.github.io/} \cite{lim2019korquad1}}\footnote{\url{http://www.aihub.or.kr/aidata/84}} and sentiment analysis\footnote{\url{https://github.com/e9t/nsmc}}, but not NLI or STS.
We believe that the lack of publicly available benchmark datasets for Korean NLI and STS has led to the lack of interest for building Korean NLU models suited for these key understanding tasks.

Motivated by this, we construct and release \textbf{KorNLI} and \textbf{KorSTS}, two new benchmark datasets for NLI and STS in the Korean language. 
Following previous work \citep{conneau2018xnli}, we construct our datasets by machine-translating existing English training sets and by translating English development and test sets via human translators.
We then establish baselines for both KorNLI and KorSTS to facilitate research on Korean NLU.

\begin{figure*}[ht]
    \centering
    \includegraphics[width=0.8\linewidth]{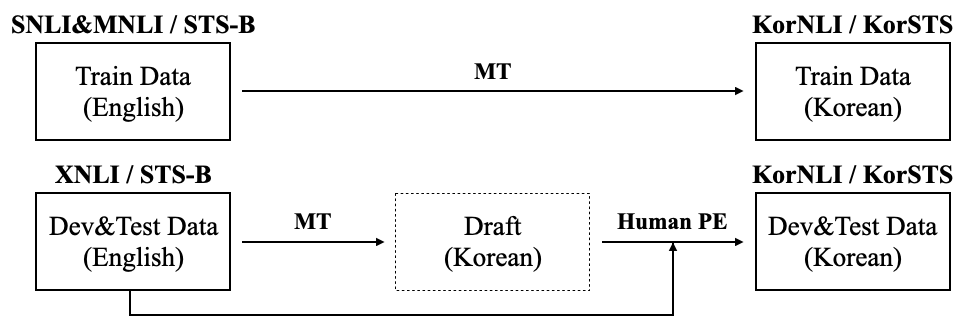}
    \caption{Data construction process. 
    MT and PE indicate machine translation and post-editing, respectively.
    We translate original English data into Korean using an internal translation engine.
    For development and test data, the machine translation outputs are further post-edited by human experts.
    }
    \label{fig:data_construct}
\end{figure*}

\section{Background}

\subsection{NLI and the \{S,M,X\}NLI Datasets}

In an NLI task, a system receives a pair of sentences, a premise and a hypothesis, and classifies their relationship into one out of three categories: \textit{entailment}, \textit{contradiction}, and \textit{neutral}.

There are several publicly available NLI datasets in English.
\citet{bowman2015large} introduced the Stanford NLI (SNLI) dataset, which consists of 570K English sentence pairs based on image captions.
\citet{williams2018broad} introduced the Multi-Genre NLI (MNLI) dataset, which consists of 455K English sentence pairs from ten genres. 
\citet{conneau2018xnli} released the Cross-lingual NLI (XNLI) dataset by extending the development and test data of the MNLI corpus to 15 languages.
Note that Korean is not one of the 15 languages in XNLI.
There are also publicly available NLI datasets in a few other non-English languages \citep{fonseca2016assin,real2019organizing,hayashibe2020japanese}, but none exists for Korean at the time of publication.

\subsection{STS and the STS-B Dataset}

STS is a task that assesses the gradations of semantic similarity between two sentences. 
The similarity score ranges from 0 (completely dissimilar) to 5 (completely equivalent). 
It is commonly used to evaluate either how well a model grasps the closeness of two sentences in meaning, or how well a sentence embedding embodies the semantic representation of the sentence. 

The STS-B dataset consists of 8,628 English sentence pairs selected from the STS tasks organized in the context of SemEval between 2012 and 2017~\citep{agirre2012semeval,agirre2013sem,agirre2014semeval,agirre2015semeval,agirre2016semeval}. 
The domain of input sentences covers image captions, news headlines, and user forums.
For details, we refer readers to \citet{cer2017semeval}.

\section{Data}

\subsection{Data Construction}

We explain how we develop two new Korean language understanding datasets: KorNLI and KorSTS.
The KorNLI dataset is derived from three different sources: SNLI, MNLI, and XNLI, while the KorSTS dataset stems from the STS-B dataset. The overall construction process, which is applied identically to the two new datasets, is illustrated in Figure~\ref{fig:data_construct}. 

First, we translate the training sets of the SNLI, MNLI, and STS-B datasets, as well as the development and test sets of the XNLI\footnote{Only English examples count.} and STS-B datasets, into Korean using an internal neural machine translation engine.
Then, the translation results of the development and test sets are post-edited by professional translators in order to guarantee the quality of evaluation. 
This multi-stage translation strategy aims not only to expedite the translators' work, but also to help maintain the translation consistency between the training and evaluation datasets. 
It is worth noting that the post-editing procedure does not simply mean proofreading. 
Rather, it refers to human translation based on the prior machine translation results, which serve as first drafts.

\subsubsection{Translation Quality}

To ensure translation quality, we hired two professional translators with at least seven years of experience who specialize in academic papers/books as well as business contracts. 
The two translators each post-edited half of the dataset and cross-checked each other’s translation afterward. This was further examined by one of the authors, who is fluent in both English and Korean.

We also note that the professional translators did not have to edit much during post-editing, suggesting that the machine-translated sentences were often good enough to begin with.
We found that the BLEU scores between the machine-translated and post-edited sentences were 63.30 for KorNLI and 73.26 for KorSTS, and for approximately half the time (47\% for KorNLI and 53\% for KorSTS), the translators did not have to change the machine-translated sentence at all.

Finally, we note that translators did not see the English gold labels during post-editing, in order to expedite the post-editing process. 
See Section \ref{sec:dataquality} for a discussion on the effect of translation on data quality.

\setlength{\tabcolsep}{4pt}
\subsection{KorNLI}
\begin{table}[t]
    \small
    \centering
    \begin{tabular}{c|cccc}
    \Xhline{1.1pt}
    \bf KorNLI & \bf Total & \bf Train & \bf Dev. & \bf Test \\
    \hline
    Source & - & SNLI, MNLI & XNLI & XNLI \\
    Translated by & - & Machine & Human & Human \\
    \# Examples & 950,354 & 942,854 & 2,490 & 5,010 \\
     \# Words (P) &  13.6 & 13.6 & 13.0 & 13.1 \\
     \# Words (H) & 7.1 & 7.2 & 6.8 & 6.8 \\
    \Xhline{1.1pt}
    \end{tabular}
    \caption{Statistics of KorNLI dataset. The last two rows mean the average number of words in a Premise (P) and a Hypothesis (H), respectively.}
    \label{tab:stat-kornli}
\end{table}

\setlength{\tabcolsep}{4pt}
\begin{table}[t]
\small
\centering
\begin{tabular}{l|c}
\Xhline{1.1pt}
\bf Examples & \bf Label \\
\hline
\textbf{P}: \begin{CJK}{UTF8}{mj}너는 거기에 있을 필요 없어.\end{CJK} & \multirow{4}{*}{E} \\
\hspace{9pt} ``You don’t have to stay there.'' & \\
\textbf{H}: \begin{CJK}{UTF8}{mj}가도 돼.\end{CJK} & \\
\hspace{9pt} ``You can leave.'' & \\
\hline
\textbf{P}: \begin{CJK}{UTF8}{mj}너는 거기에 있을 필요 없어.\end{CJK} & \multirow{4}{*}{C} \\
\hspace{9pt} ``You don’t have to stay there.'' & \\
\textbf{H}: \begin{CJK}{UTF8}{mj}넌 정확히 그 자리에 있어야 해! \end{CJK} & \\
\hspace{9pt} ``You need to stay in this place exactly!'' & \\
\hline
\textbf{P}: \begin{CJK}{UTF8}{mj}너는 거기에 있을 필요 없어.\end{CJK} & \multirow{4}{*}{N} \\
\hspace{9pt} ``You don’t have to stay there.'' & \\
\textbf{H}: \begin{CJK}{UTF8}{mj}네가 원하면 넌 집에 가도 돼.\end{CJK} & \\
\hspace{9pt} ``You can go home if you like.'' & \\
\Xhline{1.1pt}
\end{tabular}

\caption{Examples from KorNLI dataset. \textbf{P}: Premise, \textbf{H}: Hypothesis. E: Entailment, C: Contradiction, N: Neutral.}
\label{tab:kornli-example}
\end{table}

Table \ref{tab:stat-kornli} shows the statistics of the KorNLI dataset. There are 942,854 training examples translated automatically and 7,500 evaluation (development and test) examples translated manually. The premises are almost twice as long as the hypotheses, as reported in \citet{conneau2018xnli}.
We present a few examples in Table \ref{tab:kornli-example}.

\subsection{KorSTS}

\setlength{\tabcolsep}{4pt}
\begin{table}[t]
\small
\begin{center}

\begin{tabular}{c|cccc}
\Xhline{1.1pt}
    \bf KorSTS & \bf Total & \bf Train & \bf Dev. & \bf Test \\
    \hline
    Source & - & STS-B & STS-B & STS-B \\
    Translated by & - & Machine & Human & Human \\
    \# Examples & 8,628 & 5,749 & 1,500 & 1,379 \\
    Avg. \# Words & 7.7 & 7.5 & 8.7 & 7.6 \\
\Xhline{1.1pt}
\end{tabular}

\end{center}
\caption{\label{tab:stat-korsts} Statistics of KorSTS dataset. 
}
\end{table}

\setlength{\tabcolsep}{4pt}
\begin{table}[t]
\small
\centering
\begin{tabular}{l|c}
\Xhline{1.1pt}
\bf Examples & \bf Score \\
\hline
\begin{CJK}{UTF8}{mj}\textbf{A}: 한 남자가 음식을 먹고 있다.\end{CJK} & \multirow{4}{*}{4.2} \\ 
\hspace{9pt} ``A man is eating food.'' & \\
\textbf{B}: \begin{CJK}{UTF8}{mj}한 남자가 뭔가를 먹고 있다.\end{CJK} & \\
\hspace{9pt} ``A man is eating something.'' & \\
\hline
\begin{CJK}{UTF8}{mj}\textbf{A}: 한 여성이 고기를 요리하고 있다.\end{CJK} & \multirow{4}{*}{0.0} \\
\hspace{9pt} ``A woman is cooking meat.'' & \\
\textbf{B}: \begin{CJK}{UTF8}{mj}한 남자가 말하고 있다.\end{CJK} & \\
\hspace{9pt} ``A man is speaking.'' & \\
\Xhline{1.1pt}
\end{tabular}

\caption{Examples from KorSTS dataset.}
\label{tab:korsts-example}
\end{table}

As provided in Table \ref{tab:stat-korsts}, the KorSTS dataset comprises 5,749 training examples translated automatically and 2,879 evaluation examples translated manually. 
Examples are shown in Table \ref{tab:korsts-example}. 

\section{Baselines}\label{sec:baselines}

In this section, we provide baselines for the Korean NLI and STS tasks using our newly created benchmark datasets.
Because both tasks receive a pair of sentences as an input, there are two different approaches depending on whether the model encodes the sentences jointly (``cross-encoding'') or separately (``bi-encoding'').\footnote{These nomenclatures (cross-encoding and bi-encoding) are adopted from \citet{humeau2020poly}.}

\subsection{Cross-encoding Approaches}\label{sec:cross}

\setlength{\tabcolsep}{2pt}
\begin{table}[htb]
\small
\begin{center}
\begin{tabular}{lc|cc}
     \Xhline{1.1pt}
     \bf Model & \bf \# Params. & \bf \textsuperscript{\textdagger}KorNLI & \bf KorSTS \\
     \hline
     \multicolumn{4}{l}{\it Fine-tuned on Korean training set} \\ \hline
     Korean RoBERTa (base) & 111M & 82.75 & 83.00 \\
     Korean RoBERTa (large) & 338M & \bf 83.67 & \bf 85.27 \\
     XLM-R (base) & 270M & 80.56 & 77.78 \\
     XLM-R (large) & 550M & 83.41 & 84.68 \\ \hline
     \multicolumn{4}{l}{\it Fine-tuned on English training set (Cross-lingual Transfer)} \\ \hline
     XLM-R (base) & 270M & 75.17 & - \\
     XLM-R (large) & 550M & 80.30 & - \\ 
     \Xhline{1.1pt}
\end{tabular}
\caption{KorNLI and KorSTS test set scores for fine-tuned \emph{cross-encoding} language models. KorNLI scores are accuracy (\%) and KorSTS scores are 100 $\times$ Spearman correlation. \textsuperscript{\textdagger}To ensure comparability with XNLI, we only use the MNLI portion of the KorNLI dataset.}
\label{tab:baseline_fine-tuned}
\end{center}
\end{table}

\setlength{\tabcolsep}{4pt}
\begin{table*}[tb]
\small
\begin{center}
\begin{tabular}{lc|cc|cc}
    \Xhline{1.1pt}
    \multirow{5}{*}{\bf Model} & \multirow{5}{*}{\bf \# Params.} & \multicolumn{4}{c}{\bf KorSTS} \\
    \cline{3-6}
     & & \multicolumn{2}{c|}{\bf Unsupervised} & \multicolumn{2}{c}{\bf Supervised} \\
     \cline{3-6}
     &  & \it - & \thead{\it Trained on: \\ \it KorNLI} & \thead{\it Trained on: \\ \it KorSTS} & \thead{\it Trained on: \\ \it KorNLI \\ $\rightarrow$ \it KorSTS} \\
    \hline
    Korean fastText  & - & 47.96 & - & - & -  \\
    M-USE\textsubscript{CNN} (base) & 68.9M & - & \textsuperscript{\textdagger}72.74 & - & -\\
    M-USE\textsubscript{CNN} (large) & 85.2M & - & \textsuperscript{\textdagger}76.32 & - & - \\ \hline
    Korean SRoBERTa (base) & 111M & 48.96 & 74.19 & 78.94 & 80.29  \\
    Korean SRoBERTa (large) & 338M &  51.35 & 75.46 & \bf 79.55 & 80.49 \\
    SXLM-R  (base) & 270M & 45.05 & 73.99 & 68.36 & 79.13 \\
    SXLM-R  (large) & 550M & 39.92 & \bf 77.01 & 77.71 & \bf 81.84 \\
    \Xhline{1.1pt}
\end{tabular}
\caption{KorSTS test set scores (100 $\times$ Spearman correlation) of \emph{bi-encoding} models. Note that the first two columns of results are unsupervised w.r.t. KorSTS, and the latter two are supervised w.r.t. KorSTS. \textsuperscript{\textdagger}Trained on machine-translated SNLI only.}
\label{tab:baseline_sentence}
\end{center}
\end{table*}

As illustrated with BERT \cite{devlin2018bert} and many of its variants, the \textit{de facto} standard approach for NLU tasks is to pre-train a large language model and fine-tune it on each task.
In the cross-encoding approach, the pre-trained language model takes each sentence pair as a single input for fine-tuning.
These cross-encoding models typically achieve the state-of-the-art performance over bi-encoding models, which encode each input sentence separately. 

For both KorNLI and KorSTS, we consider two pre-trained language models. 
We first pre-train a Korean RoBERTa \cite{liu2019roberta}, both base and large versions, on a collection of internally collected Korean corpora (65GB). 
We construct a byte pair encoding (BPE)~\cite{gage1994new,sennrich2016neural} dictionary of 32K tokens using SentencePiece \cite{kudo2018sentencepiece}.
We train our models using \texttt{fairseq} \cite{ott2019fairseq} with 32 V100 GPUs for the base model (25 days) and 64 for the large model (20 days). 

We also use XLM-R \cite{conneau2019cross}, a publicly available cross-lingual language model that was pre-trained on 2.5TB of Common Crawl corpora in 100 languages including Korean (54GB).
Note that the base and large architectures of XLM-R are identical to those of RoBERTa, except that the vocabulary size is significantly larger (250K), making the embedding and output layers that much larger.

In Table~\ref{tab:baseline_fine-tuned}, we report the test set scores for cross-encoding models fine-tuned on KorNLI (accuracy) and KorSTS (Spearman correlation). 
For KorNLI, we additionally include results for XLM-R models fine-tuned on the original MNLI training set (also known as \emph{cross-lingual transfer} in XNLI).
To ensure comparability across settings, we only train on the MNLI portion when fine-tuning on KorNLI.

Overall, the Korean RoBERTa models outperform the XLM-R models, regardless of whether they are fine-tuned on Korean or English training sets.
For each model, the larger variant outperforms the base one, consistent with previous findings.
The large version of Korean RoBERTa performs the best for both KorNLI (83.67\%) and KorSTS (85.27\%) among all models tested.
Among the XLM-R models for KorNLI, those fine-tuned on the Korean training set consistently outperform the cross-lingual transfer variants.

\subsection{Bi-encoding Approaches}\label{sec:bi}

We also report the KorSTS scores of bi-encoding models. 
The bi-encoding approach bears practical importance in applications such as semantic search, where computing pairwise similarity among a large set of sentences is computationally expensive with cross-encoding.

Here, we first provide two baselines that do not use pre-trained language models: Korean fastText and the multilingual universal sentence encoder (M-USE). 
Korean fastText \cite{bojanowski2017enriching} is a pre-trained word embedding model\footnote{\url{https://dl.fbaipublicfiles.com/fasttext/vectors-crawl/cc.ko.300.bin.gz}} trained on Korean text from Common Crawl. 
To produce sentence embeddings, we take the average of fastText word embeddings for each sentence.
M-USE\footnote{\url{https://tfhub.dev/google/universal-sentence-encoder-multilingual/3}}~\cite{yang2019multilingual}, is a CNN-based sentence encoder model trained for NLI, question-answering, and translation ranking across 16 languages including Korean. 
For both Korean fastText and M-USE, we compute the cosine similarity between two input sentence embeddings to make an unsupervised STS prediction.

Pre-trained language models can also be used as bi-encoding models following the approach of SentenceBERT~\cite{reimers2019sentencebert}, which involves fine-tuning a BERT-like model with a Siamese network structure on NLI and/or STS. 
We use the SentenceBERT approach for both Korean RoBERTa (``Korean SRoBERTa'') and XLM-R (``SXLM-R'').
We adopt the \texttt{MEAN} pooling strategy, i.e., computing the sentence vector as the mean of all contextualized word vectors.

In Table~\ref{tab:baseline_sentence}, we present the KorSTS test set scores (100 $\times$ Spearman correlation) for the bi-encoding models.
We categorize each result based on whether the model was additionally trained on KorNLI and/or KorSTS. 
Note that models that are not fine-tuned at all or only fine-tuned to KorNLI can be considered as unsupervised w.r.t. KorSTS.
Also note that M-USE is trained on a machine-translated version of SNLI, which is a subset of KorNLI, as part of its pre-training step.

First, given each model, we find that supplementary training on KorNLI consistently improves the KorSTS scores for both unsupervised and supervised settings, as was the case with English models \cite{conneau2017supervised,reimers2019sentencebert}.
This shows that the KorNLI dataset can be an effective intermediate training source for bi-encoding approaches.
When comparing the baseline models in each setting, we find that both M-USE and the SentenceBERT-based models trained on KorNLI achieve competitive unsupervised KorSTS scores.
Both models significantly outperform the average of fastText embeddings model and the Korean SRoBERTa and SXLM-R models without fine-tuning.
Among our baselines, large SXLM-R trained on KorNLI followed by KorSTS achieves the best score (81.84).

\section{Effect of Translation on Data Quality}\label{sec:dataquality}

As noted in \citep{conneau2018xnli}, translation quality does not necessarily guarantee that the semantic relationships between sentences are preserved. 
We also translated each sentence independently and took the gold labels from the original English pair, so the resulting label might no longer be ``gold,'' due to both incorrect translations and (in rarer cases) linguistic differences that make it difficult to translate specific concepts. 

Fortunately, it was also pointed out in \citep{conneau2018xnli} that annotators could recover the NLI labels at a similar accuracy in translated pairs (83\% in French) as in original pairs (85\% in English). 
In addition, our baseline experiments in Section \ref{sec:cross} show that supplementary training on KorNLI improves KorSTS performance (+1\% for RoBERTa and +4-11\% for XLM-R), suggesting that the labels of KorNLI are still meaningful. 
Another quantitative evidence is that the performance of XLM-R fine-tuned on KorNLI (80.3\% with cross-lingual transfer) is within a comparable range of the model’s performance on other XNLI languages (80.1\% on average). 

Nevertheless, we could also find some (not many) examples the gold label becomes incorrect after translating input sentences to Korean. 
For example, there were cases in which the two input sentences for KorSTS were so similar (with 4+ similarity scores) that upon translation, the two inputs simply became identical. 
In another case, the English word \textit{sir} appeared in the premise of an NLI example and was translated to \begin{CJK}{UTF8}{mj}{선생님}\end{CJK}, which is a correct word translation but is a gender-neutral noun, because there is no gender-specific counterpart to the word in Korean. 
As a result, when the hypothesis referencing the entity as \textit{the man} got translated into \begin{CJK}{UTF8}{mj}{남자}\end{CJK} (gender-specific), the English gold label (entailment) was no longer correct in the translated example. 
More systematically analyzing these errors is an interesting future work, although the amount of human efforts involved in this analysis would match that of labeling a new dataset.

\section{Conclusion}
We introduced KorNLI and KorSTS---new datasets for Korean natural language understanding. 
Using these datasets, we also established baselines for Korean NLI and STS with both cross-encoding and bi-encoding approaches.
Looking forward, we hope that our datasets and baselines will facilitate future research on not only improving Korean NLU systems but also increasing language diversity in NLU research.

\section*{Acknowledgements}

We thank Pulip Park for helping with hiring and contacting with the professional translators. 
We would also like to acknowledge Kakao Brain Cloud, which we used for our baseline experiments.


\bibliography{anthology,emnlp2020}
\bibliographystyle{acl_natbib}

\appendix

\section{Korean RoBERTa Pre-training}
\label{app:pretraining}

For the Korean RoBERTa baselines used in \S\ref{sec:baselines}, we pre-train a RoBERTa \citep{liu2019roberta} model on an internal Korean corpora of size 65GB, consisting of online news articles (56GB), encyclopedia (7GB), movie subtitles ($\sim$1GB), and the Sejong corpus\footnote{\url{https://ithub.korean.go.kr/user/guide/corpus/guide1.do}} ($\sim$0.5GB).
We use \texttt{fairseq} \citep{ott2019fairseq}, which includes the official implementation for RoBERTa.

\setlength{\tabcolsep}{4pt}
\begin{table}[ht]
\small
    \centering
    \begin{tabular}{l|c|c}
        \Xhline{1.1pt}
         \bf Hyperparameter    & \bf Large         & \bf Base         \\ \hline
         Total \# of Parameters& 338M              & 111M             \\ \hline 
         Number of Layers      & 24                & 12               \\
         Hidden Size           & 1024              & 768              \\
         FFN Inner Hidden Size & 4096              & 3072             \\
         Attention Heads       & 16                & 12               \\
         Attention Head Size   & 64                & 64               \\
         Dropout               & 0.1               & 0.1              \\
         Attention Dropout     & 0.1               & 0.1              \\
         Warmup Steps          & 30K               & 24K              \\
         Peak Learning Rate    & 2e-4              & 6e-4             \\
         Batch Size            & 2048              & 8192             \\
         Weight Decay          & 0.01              & 0.01             \\
         Scheduled \# Updates  & 2M                & 500K             \\
         Performed \# Updates\textsuperscript{*}   & 502.3K            & 500K             \\
         Learning Rate Decay   & Linear            & Linear           \\
         Adam $\epsilon$       & 1e-6              & 1e-6             \\
         Adam $\beta_1$        & 0.9               & 0.9              \\
         Adam $\beta_2$        & 0.98              & 0.98             \\
         Gradient Clipping     & 0.0               & 0.0              \\ 
         \Xhline{1.1pt}
    \end{tabular}
    \caption{Hyperparameters for Korean RoBERTa pre-training. \textsuperscript{*}For the large model, we initially scheduled our learning rate to decay to zero at 2M steps. After 500K steps, however, we observed no significant improvement in the KorNLI and KorSTS fine-tuning performance.}
    \label{tab:hparams_pretraining}
\end{table}

In Table \ref{tab:hparams_pretraining}, we list all hyperparameters we use for Korean RoBERTa pre-training.
Note that, compared to the original RoBERTa (English), the model architectures are identical except for the token embedding layer, as we use different vocabularies (32K \texttt{sentencepiece} vocab instead of 50K byte-level BPE).
After training, the base and large models achieve validation perplexities of 2.55 and 2.39 respectively, where the validation set is a random 5\% subset of the entire corpora.

\section{Fine-tuning with Cross-encoding Approaches}
\label{app:cross}

To fine-tune Korean RoBERTa and XLM-R models using the cross-encoding approach (\S\ref{sec:cross}), we follow the fine-tuning procedures of RoBERTa \cite{liu2019roberta} on MNLI and STS-B, as described in RoBERTa's code release\footnote{\url{https://github.com/pytorch/fairseq/blob/v0.9.0/examples/roberta/README.glue.md}}.

\setlength{\tabcolsep}{4pt}
\begin{table}[ht]
\small
    \centering
    \begin{tabular}{l|c|c}
        \Xhline{1.1pt}
         \bf Hyperparameter & \bf KorNLI & \bf KorSTS \\ \hline
         Batch Size & 32 & 16 \\
         Learning Rate Schedule & Linear & Linear \\
         Peak Learning Rate & 1e-5 & 2e-5 \\
         \# Warmup Steps & 7318 & 214 \\
         Total \# Updates & 121979 & 3596 \\
         \Xhline{1.1pt}
    \end{tabular}
    \caption{Hyperparameters for Korean RoBERTa and XLM-R fine-tuning using the \emph{cross-encoding} approach.}
    \label{tab:hparams_finetuning}
\end{table}

The fine-tuning hyperparameters are summarized in Table \ref{tab:hparams_finetuning}.
For each dataset and model size, we choose the hyperparameter configurations that are used in the corresponding English version of the dataset and model size (except for the XLM-R cross-lingual transfer using MNLI, where we also use the same hyperparameters as RoBERTa and XLM-R on KorNLI). 
We find that the hyperparameters used for English models and datasets give sufficiently good performances on the development set, so we do not perform an additional hyperparameter search.
After training each model for 10 epochs, we choose the model checkpoint that achieve the highest score on the development set and evaluate it on the test set to obtain our final results in \S\ref{sec:cross}.

We also report the development set scores for the best checkpoint in Table \ref{tab:cross_dev}.
We observe that the XLM-R models fine-tuned on KorNLI and KorSTS achieve the highest scores on the development set, although the Korean RoBERTa models perform better on the test set (Table \ref{tab:baseline_fine-tuned} in \S\ref{sec:cross}). 
Both models outperform the cross-lingual transfer models on the development set, as is the case on the test set.

\setlength{\tabcolsep}{2pt}
\begin{table}[htb]
\small
\begin{center}
\begin{tabular}{lc|cc}
     \Xhline{1.1pt}
     \bf Model & \bf \# Params. & \bf \textsuperscript{\textdagger}KorNLI & \bf KorSTS \\
     \hline
     \multicolumn{4}{l}{\it Fine-tuned on Korean training set} \\ \hline
     Korean RoBERTa (base) & 111M & 81.97 & 84.97 \\
     Korean RoBERTa (large) & 338M & 83.17 & 87.82 \\
     XLM-R (base) & 270M & 79.20 & 83.02 \\
     XLM-R (large) & 550M & \bf 84.42 & \bf 88.37 \\ \hline
     \multicolumn{4}{l}{\it Fine-tuned on English training set (Cross-lingual Transfer)} \\ \hline
     XLM-R (base) & 270M & 74.34 & - \\
     XLM-R (large) & 550M & 81.45 & - \\ 
     \Xhline{1.1pt}
\end{tabular}
\caption{KorNLI and KorSTS \textbf{development} set scores for fine-tuned \emph{cross-encoding} language models. KorNLI scores are accuracy (\%) and KorSTS scores are 100 $\times$ Spearman correlation. \textsuperscript{\textdagger}To ensure comparability with XNLI, we only use the MNLI portion of the KorNLI dataset.}
\label{tab:cross_dev}
\end{center}
\end{table}

\section{Fine-tuning with Bi-encoding Approaches}
\label{app:bi}

To fine-tune Korean RoBERTa and XLM-R models using the bi-encoding approach (\S\ref{sec:bi}), we train Korean Sentence RoBERTa (``Korean SRoBERTa'') and Sentence XLM-R (``SXLM-R''), following the fine-tuning procedure of SentenceBERT \citep{reimers2019sentencebert}.

Unless described otherwise, we follow the experimental settings, including all hyperparameters, of SentenceBERT\footnote{\url{https://github.com/UKPLab/sentence-transformers}}. 
For each model size, we manually search among learning rates \{2e-5, 1e-5\} for training on KorNLI, \{1e-5, 2e-6\} for training on KorSTS, and \{1e-5, 2e-6\} for training on KorSTS after KorNLI. 
After training until convergence, we choose the learning rate that lead to the highest KorSTS score on the development set. 
These hyperparameters are shown in Table \ref{tab:hparams_finetuning_biencoding}.

\setlength{\tabcolsep}{2pt}
\begin{table}[htb]
    \small
    \centering
    \begin{tabular}{l|c|c|c}
        \Xhline{1.1pt}
         \multirow{3}{*}{\bf Model} & \multirow{3}{*}{\bf KorNLI} & \multirow{3}{*}{\bf KorSTS} &\bf KorSTS \\ 
         & & & \bf (after \\ 
         & & & \bf KorNLI) \\ \hline
        Korean SRoBERTa (base) & 2e-5 & 1e-5 & 1e-5 \\
        Korean SRoBERTa (large) & 2e-5 & 1e-5 & 1e-5 \\
        SXLM-R (base) & 2e-5 & 1e-5 & 1e-5 \\
        SXLM-R (large) & 1e-5 & 2e-6 & 1e-5 \\
         \Xhline{1.1pt}
    \end{tabular}
    \caption{Learning rates for Korean SRoBERTa and SXLM-R fine-tuning using the \emph{bi-encoding} approach.}
    \label{tab:hparams_finetuning_biencoding}
\end{table}

\setlength{\tabcolsep}{4pt}
\begin{table*}[htb]
\small
\begin{center}
\begin{tabular}{lc|cc|cc}
    \Xhline{1.1pt}
    \multirow{5}{*}{\bf Model} & \multirow{5}{*}{\bf \# Params.} & \multicolumn{4}{c}{\bf KorSTS} \\
    \cline{3-6}
     & & \multicolumn{2}{c|}{\bf Unsupervised} & \multicolumn{2}{c}{\bf Supervised} \\
     \cline{3-6}
     &  & \it - & \thead{\it Trained on: \\ \it KorNLI} & \thead{\it Trained on: \\ \it KorSTS} & \thead{\it Trained on: \\ \it KorNLI \\ $\rightarrow$ \it KorSTS} \\
    \hline
    
    Korean SRoBERTa (base) & 111M & 63.34 & 76.48 & 83.68 & 83.54  \\
    Korean SRoBERTa (large) & 338M &  60.15 & 77.95 & \bf 84.74 & \bf 84.21 \\
    SXLM-R  (base) & 270M & 64.27 & 77.65 & 74.60 & 81.95 \\
    SXLM-R  (large) & 550M & 55.00 & \bf 79.16 & 82.66 &  84.13 \\
    \Xhline{1.1pt}
\end{tabular}
\caption{KorSTS \textbf{development} set scores (100 $\times$ Spearman correlation) of \emph{bi-encoding} models. Note that the first two columns of results are unsupervised w.r.t. KorSTS, and the latter two are supervised w.r.t. KorSTS.}
\label{tab:bi_dev}
\end{center}
\end{table*}

We report the development set scores in Table \ref{tab:bi_dev}. 
Korean SRoBERTa (large) achieves the best development set performance on both supervised settings, but SXLM-R (large) achieves the best performance for the \textit{KorNLI $\rightarrow$ KorSTS} setting on test set.

\end{document}